# Detail-aware multi-view stereo network for depth estimation

HAITAO TIAN,[1] JUNYANG LI,[1] CHENXING WANG,[1,*] HELONG JIANG,[1]

[1]School of Automation, Southeast University, Nanjing 210000, China
*cxwang@seu.edu.cn

**Abstract：**
**Multi-view stereo methods have achieved great success for depth estimation based on the coarse-to-fine depth learning frameworks, however, the existing methods perform poorly in recovering the depth of object boundaries and detail regions. To address these issues, we propose a detail-aware multi-view stereo network (DA-MVSNet) with a coarse-to-fine framework. The geometric depth clues hidden in the coarse stage are utilized to maintain the geometric structural relationships between object surfaces and enhance the expressive capability of image features. In addition, an image synthesis loss is employed to constrain the gradient flow for detailed regions and further strengthen the supervision of object boundaries and texture-rich areas. Finally, we propose an adaptive depth interval adjustment strategy to improve the accuracy of object reconstruction. Extensive experiments on the DTU and Tanks & Temples datasets demonstrate that our method achieves competitive results. The code is available at https://github.com/wsmtht520-/DAMVSNet.**

## 1. Introduction

The Multi-view stereo (MVS) technique can reconstruct a 3D scene from multiple calibrated images according to the matching relationship and stereo correspondences between multi-view images. To achieve promising reconstruction results, the traditional MVS-based [1,2] and PatchMatch-based methods [3,4] require texture-rich and restricted lighting conditions. Alternatively, the deep learning-based approaches [5-8] try to take advantage of global scene semantic information, including environmental illumination and object materials, to maintain high performance in complex lighting. The key to these methods is to warp deep image features into the reference camera frustum so that the 3D cost volume can be built via a differentiable homography with sampled depth hypotheses. Then, the 3D cost volumes are processed by 3D CNNs to regress a depth map of the reference image.

Though promising results have been achieved, the learning-based MVS methods remain challenging in the depth retrieval on the edge or detailed regions. One of the causes is that these methods did not take advantage of the geometric clues embedded in the MVS scenarios sufficiently; another cause lies in the overlearning of the larger proportion of hard-sample regions sometimes, e.g., texture less regions, while the small detailed areas, such as the edge regions, are learned insufficiently.

To tackle these issues, we design a geometric depth embedding (GDE) module to explicitly integrate the geometric depth in the coarse stage, with the features extracted by the classic Feature Pyramid Network (FPN) [9]. Meanwhile, we design an image synthesis loss (IS Loss) to supervise the depth values by transforming them into a textured image that can describe the detailed information more sufficiently. Besides, we propose an adaptive depth interval adjustment (ADIA) strategy to allocate more effective depth hypothesis planes close to the potential GT depth value of the object surfaces, while previous methods, such as CasMVSNet [10] and UCSNet [11], adopt equal depth hypothesis planes to construct the 3D cost volume.

In summary, our contributions are listed as follows:

1、We propose to embed the geometric depth clues hidden in the coarse stages into the image features of the finer stages to preserve the geometric structural relationships between object surfaces and ensure robust cost matching.

2、We design an image synthesis Loss to enhance depth supervision for details during training.

3、We propose an ADIA strategy to achieve adaptive variable depth interval partition to estimate more accurate depth values.

The above modules are plug-and-play and can be easily integrated into the MVSNet-based methods. Experimental results on the challenging datasets show that our methods display competitive results compared with SOTA methods.

## 2. Related Works

### 2.1 Traditional Multi-View Stereo Methods

The traditional MVS methods make use of various 3D representations, such as mesh [12], point cloud [13], voxel [14,15], and depth map [16,17]. Among these, the depth-based methods can obtain a more complete surface reconstruction with higher robustness. They avoided solving the intractable topology problem by formulating the multi-view reconstruction into a depth estimation problem and fusing all depth maps to form a single 3D point cloud. Among them, COLMAP [2] and ACMM [18] can obtain more stable results. COLMAP estimates the pixel-wise depth and normal value using photometric and geometric priors. ACMM employs multi-scale geometric consistency to reconstruct features at different scales. However, if the scenarios are complicated, these traditional MVS methods may produce obvious artifacts due to large noise and poor correspondences. Moreover, traditional pipelines rely on hand-crafted opera-tors for feature matching, resulting in poor stability of image matching.

## 2.2 Learning-based Multi-View Stereo

In recent years, many approaches estimated depth maps utilizing deep learning techniques. SurfaceNet [7] and LSM [19] are the first MVS methods based on volumetric learning to regress surface voxels from 3D space. However, they are restricted by the large memory requirement. MVSNet [8] first realized an end-to-end pipeline with saved memory, which builds 3D cost volume to aggregate the warped features from the reference and source images and then outputs a depth map by a 3D CNN. The MVSNet reduces some procedure steps to improve efficiency, but the dense hypotheses planes and 3D cost volumes still require a heavy memory load. Some strategies are proposed to reduce the potential capacity of the MVSNet pipeline, such as the recurrent MVSNet architectures [20,21], coarse-to-fine manner [10], and multi-stage binary search [22], and other ones are later explored to increase the depth quality further, such as coarse-to-fine depth optimization [23-26], and patch matching-based method [27]. Furthermore, PVSNet [28], AA-RMVSNet [29], and Vis-MVSNet [30] employed pixel-wise visibility mechanisms to aggregate reliable matching costs to solve the occlusion issues.

However, the depth maps reconstructed from MVS networks still display low quality in some challenging regions with illumination changes, repetitive texture, or fewer textures. The Transformer [31] modules have advantages in extracting long-range global context using self-attention and cross-attention blocks. Motivated by this, TransMVSNet [32] first introduces the transformers to enhance feature representation in the challenging areas. Some other methods also introduce transformers [33-39] to improve the accuracy significantly but still at the cost of heavy network parameters. Then, LE-MVSNet [40] and LS-MVSNet [41] were proposed due to their light weight. However, the depth values reconstructed in detailed regions are still unsatisfactory. For example, some deep learning methods based on cost volume may predict many outliers in the edge regions.

## 2.3 Supervision of Multi-view Stereo Network

Most of the learning-based MVS methods are supervised by $L1$ loss, such as the CasMVSNet [10], UCSNet [11], and PatchmatchNet [27], while some other methods, like CT-MVSNet [39], WT-MVSNet [40], and GBiNet [22] use the cross-entropy function as the loss. In general, using a regression manner yields higher reconstruction accuracy while lower confidence in MVS problems than using classification means, and vice versa [36]. To this end, UniMVSNet [42] unifies the advantages of regression and classification by a unified focal loss. MVSFormer [36] and MVSFormer++ [37] further optimize this by introducing a temperature adjustment coefficient that allows adaptive switching to select the appropriate loss function for classifying or regressing during different stages.

These loss functions all focus on the information in the depth domain. However, a depth map generally displays smoothly, and the training procedure focuses on learning of large smooth regions but omits the small regions containing detailed information. To solve this problem, we introduce the textured image to detect detailed regions and propose an IS Loss to enhance the supervision of details by integrating the constraints in domains of image and depth.

## 3. Methodology
### 3.1 Network Overview

Our DA-MVSNet is also a typical multi-stage coarse-to-fine framework. There are three main contributions: the GDE module, the ADIA strategy, and the IS Loss. These three parts are plug-and-play. The CasMVSNet [10] is the baseline, and our modules are added to predict the depth.

Figure 1 illustrates the overall architecture of our DA-MVSNet. Three cascaded stages complete a coarse-to-fine procedure for depth map inference. In each stage, we use the main part of MVSNet [8], which is based on a large-scale all-pixel depth range to obtain a coarse or fine depth map.

After inputting a set of calibrated unstructured images, we select one image as the reference. In contrast, the remaining images are the source images to be processed by the FPN [9] to extract multi-scale feature maps. In each stage, the 2D feature maps are warped into several depth hypothesis planes (DHPs) of the reference camera using differentiable homography, resulting in feature volumes fused to construct a cost volume. Then, the cost volume is regularized by the 3D CNN to produce a corresponding probability volume for predicting a depth map.

Our GDE and ADIA modules are added to the last two stages only. The GDE module integrates the geometric clues with the image features through the coarser depth maps in the previous stage to enhance the geometry expression ability to the image features. The ADIA strategy is used during the construction of the cost volume to adaptively adjust the depth interval for setting the DHPs. The training process is supervised by the common $L1$ loss for the probability volume and the designed IS loss for the depth. The next sections will introduce these modules in detail.

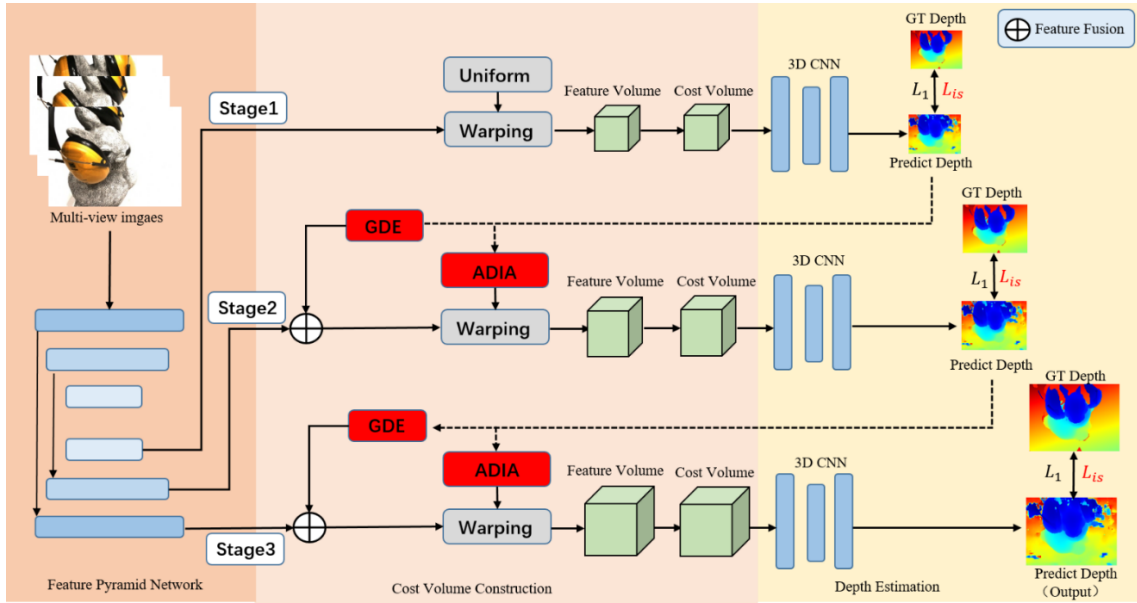

**Fig. 1.** Illustration of DA-MVSNet, with our modules marked as red.

### 3.2 Geometric Depth Embedding Module

The baseline of our method, CasMVSNet [10], can predict a relatively satisfactory depth map through a three-stage coarse-to-fine framework. However, if the scene is complex, the resulting depth map is worse, particularly at the detail regions such as the edges of objects. Therefore, we propose the GDE module that fuses the depth features with image features to improve the expression abilities for geometry and detail information. The GDE module is depicted in Fig. 2.

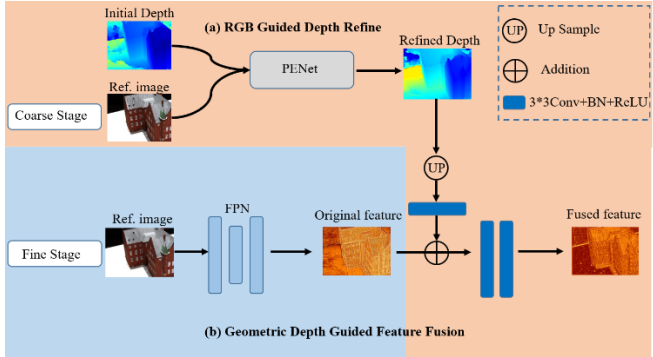

**Fig. 2.** The framework of the GDE module.

The initial depth in the upper left is the predicted depth map from the previous coarser stage. If this depth map is poor, errors may accumulate and propagate to the following stages and network training. Therefore, we introduce the RGB-guided precise and efficient network (PENet) [43] to refine the initial depth map by highlighting the contour values to improve the quality of the depth map.

The refined depth map goes through a 3×3 2D convolutional layer for extracting depth features, and then the extracted depth features are added to the image features extracted from FPN in the next finer stage. After two 2D convolutional layers, the fused features are obtained, which can describe the details sufficiently, like the image features, and embed the spatial geometry information like the depth features, thereby improving the power of feature expression abilities to cope with more complex scenes.

Figure 3 visualizes the various feature maps. The RGB feature displays complex textures that can better describe the details, while the fused feature map better combines the details and the depth.

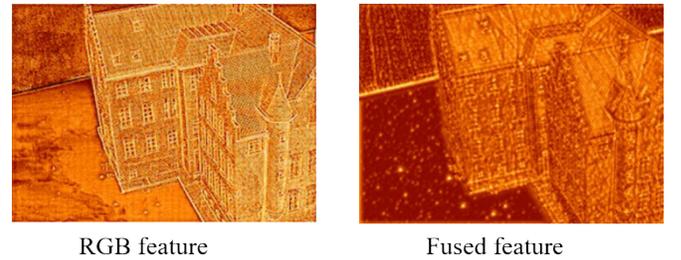

RGB feature                Fused feature

**Fig. 3.** Various types of feature maps.

### 3.3 Adaptive Depth Interval Adjustment

The key of our framework is to sub-partition the 3D cost volume progressively and refine the depth prediction with increasing resolution and accuracy. Each 3D cost volume is partitioned by $i$ DHPs, each denoted as $H_i$; correspondingly, the 3D CNN at each stage predicts a depth probability volume consisting of $i$ depth probability map still, each denoted as $P_i$ associated with $H_i$. $P_i(x)$ represents how probable the depth hypothesis at pixel $x$ is $H_i(x)$, so a depth map $\hat{D}(x)$ in one stage is reconstructed by:

$$\hat{D}(x) = \sum_{i=1}^{n} H_i(x) \cdot P_i(x), \qquad (1)$$

where $n$ is the number of DHPs.

In the above process, the depth prediction is closely related to the partitioned DHPs and the subsequently computed depth probability maps. However, the previous methods usually generate the DHPs using uniform sampling to the 3D cost volume. If the sampling number is small, the DHPs set near the potential true depth are sparse, leading to a larger distance from the determined DHP to the truth and resulting in large errors for depth prediction; if the sampling number is enlarged, the computation cost is increased. To solve this issue, we propose an ADIA strategy to allocate DHPs near the potential true depth and less away from it. To achieve this strategy, the Z-score distribution is selected [44] to measure the distance between the DHPs.

Figure 4 illustrates the principle of our ADIA strategy. The depth ranges from $d_{min}$ to $d_{max}$, and several DHPs are set within this range. The common strategy sets the DHPs with uniform intervals, as shown in the left part, but ours can adaptively set them with weighted intervals. This can make more DHPs gather around the potential true value to increase the probability of setting the correct DHP closest to the truth. Our ADIA strategy is applied to stage 2 and stage 3 only.

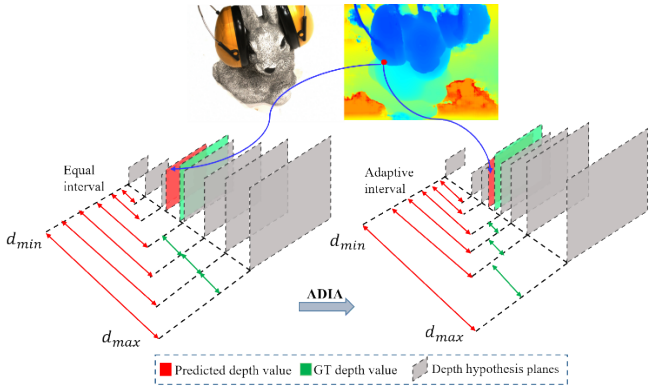

**Fig. 4.** The Adaptive Depth Interval Adjustment Module.

In stage 1, we still set the DHPs with uniform intervals. The $i^{th}$ DHP is denoted as $H_{1,i}(x)$, and the corresponding depth probability map is $P_{1,i}(x)$, so a coarse depth map $\hat{D}_1(x)$ can be predicted according to Eq. (1).

In stages 2 and 3, we adjust the uniformly set DHP $H_{k,i}(x)$ with weights, so the adaptively set DHP is obtained as:
$$\hat{H}_{k,i}(x) = H_{k,i}(x) + e_k(x) \times o_{k,i}(x), \quad (k=2, 3) \quad (2)$$
where $k$ is the stage order, $e_k(x)$ denotes the pixel-wise equal interval, and $o_{k,i}(x)$ is the offset taken as the weight.

With the depth range, $e_k(x)$ can be obtained by:
$$e_k(x) = \frac{d_k^{max}(x) - d_k^{min}(x)}{n_k}, \quad (3)$$
where $n_k$ is the empirical number of DHPs at stage $k$.

In stage 1, the depth range $[d_{min}, d_{max}]$ is pre-defined. After stage 1, we have a coarse depth map $\hat{D}_1(x)$, so the depth range for any pixel $x$ can be refined variously according to $\hat{D}_1(x)$ as $[d_{min,1}(x), d_{max,1}(x)]$. This range can be used as a priori in stage 2. Similarly, the range determined according to $\hat{D}_2(x)$ can be used as a priori in stage 3. Generally, $[d_{min,k-1}(x), d_{max,k-1}(x)]$ is used in stage $k$ ($k$=2, 3). Referring to UCSNet [11], we leverage the variance distribution for the uncertainty of $\hat{D}_{k-1}(x)$ to estimate the range $[d_{min,k-1}(x), d_{max,k-1}(x)]$. The variance $\sigma_{k-1}(x)$ of the probability distribution is calculated as
$$\sigma_{k-1}(x) = \sqrt{\sum_i^{n_{k-1}} P_{k-1,i}(x) \cdot [H_{k-1,i}(x) - \hat{D}_{k-1}(x)]^2}, \quad (4)$$

Then the pixel-wise $d_{min,k-1}(x)$ and $d_{max,k-1}(x)$ are determined as:
$$d_{min,k-1}(x) = \hat{L}_{k-1}(x) - \lambda \sigma_{k-1}(x),$$
$$d_{max,k-1}(x) = \hat{L}_{k-1}(x) + \lambda \sigma_{k-1}(x) \quad (5)$$
where $\lambda$ is a scalar factor determining how large the confidence interval is, $d_{min,k-1}(x)$ and $d_{max,k-1}(x)$ correspond to $d_k^{min}(x)$ and $d_k^{max}(x)$ in Eq. (3), respectively.

Similarly, the weight $o_{k,i}(x)$ in Eq. (2) can also be determined according to the uncertainty of the predicted depth map in the previous stage:
$$o_{k,i}(x) = softmax(\frac{H_{k,i}(x) - \hat{D}_{k-1}(x)}{\sigma_{k-1}(x)}), \quad (6)$$
where $H_{k,i}(x)$ represents the $i^{th}$ DHP of pixel $x$ at stage $k$ with equal interval, $D_{k-1}(x)$ is the predicted depth value of pixel $x$ in stage ($k$-1), and $softmax(\cdot)$ is a common activation function used in deep learning technique.

Finally, the ADIA strategy can be achieved. Note that the variance-based methods above are differentiable, enabling them to be applied in the network training.

### 3.4 Image Synthesis Loss

With the GDE and ADIA modules, the predicted depth map can be improved to higher quality after effective training. However, a large proportion of a depth map usually reveals smoothness. Hence, the supervision during the training pays more attention to the smooth regions and omits the details if only a simple common $L1$ loss is used. Therefore, we designed an image synthesis loss to magnify the gradient flows of detail regions by expressing the depth map into an RGB image.

As mentioned in subsection 3.1, the input source images have been calibrated relative to a selected reference image. With the predicted depth $\hat{D}_j$, the $j^{th}$ source image $\hat{I}_j$ can be calculated as
$$\hat{I}_j = K_j \left[ R_j \left( K_0^{-1} I_0 \hat{D}_j \right) + T_j \right], \quad (7)$$
where $I_0$ is the reference image, $K_j$ and $K_0$ denote the intrinsic camera parameters of $\hat{I}_j$ and $I_0$, respectively, $R_j$ and $T_j$ denote the rotation and translation between $\hat{I}_j$ and $I_0$, respectively.

With the transformation above, we replace the depth with the truth value to get the $j^{th}$ truth source image $I_j$. Therefore, the proposed image synthesized loss is
$$L_{IS} = \sum_{j=1}^{N} \frac{\sum \left\| (\hat{I}_j - I_j) \odot M \right\|}{m}, \quad (8)$$
where $N$ denotes the number of source images; $M$ denotes a binary mask within the valid pixels in the synthesized images, produced by ES-MVSNet [45]; $m$ is the sum number of valid pixels in the mask $M$, and $\odot$ denotes the Hadamard product. Since the texture information is more sensitive with details compared with the depth map, the designed loss $L_{IS}$ expresses the texture information synthesized into the predicted depth and can constrain the details more effectively.

The final loss for our DA-MVSNet is set by synthesizing the image domain and depth domain like

$$Loss = \sum_{a=1}^{3} (\lambda_1 L_1^a + \lambda_2 L_{IS}^a),  \quad (9)$$

where $L_1 = \|\hat{D}_j - D_j\|$ with $\hat{D}_j$ and $D_j$ denoting the ground truth and estimated depth, respectively, $\lambda_1$ and $\lambda_2$ are weight coefficients, and $a$ denotes the stage order.

## 4. Experiments

### 4.1 Datasets, Metrics, and Implementation

**Dataset.** Three datasets are used to train and test our network model:

DTU [46] is an indoor dataset consisting of 124 different objects, each scene is recorded from 49 views with 7 brightness levels. It contains ground-truth point clouds collected under well-controlled laboratory conditions.

T&T [47] dataset contains a more challenging realistic environment with large-scale variations and illumination changes. It includes an intermediate subset of 8 scenes and an advanced subset of 6.

BlendedMVS [48] dataset is a recently published large-scale synthetic dataset. It consists of over 17000 high-resolution rendered images with 3D structures.

**Metrics.** The distance metric [46,47] is used to measure the accuracy and completeness of the reconstructed point clouds. The accuracy of the reconstruction: how closely the reconstructed points lie to the ground truth. The completeness of the reconstruction evaluates to what extent all the ground-truth points are covered. Accuracy and completeness can be combined into a summary measure as the F-score, which concurrently evaluates the accuracy and completeness of the reconstruction overall.

In this study, the metric for the test on the DTU dataset is calculated referring to [46], while the one for the T&T dataset is obtained from an official website [47]. For the overall evaluation, we calculate the average of the mean accuracy and the mean completeness for DTU and use the F-score for T&T referring to [8]. Note that our metric calculations exclusively consider the foreground regions of objects, while the background regions are not included in the evaluation of the metrics.

**Implementation details.** Like the previous methods [10,33], we first train our model on the DTU training set and evaluate on the DTU evaluation set, and then finetune our model on the BlendedMVS dataset before validating the generalization of our method on T&T. As for the DTU dataset, we use 79 scenes for training, 18 scenes for validation and the rest of data for evaluation. The original image resolution is 1200×1600, and each scene has 7 lighting conditions. We crop the rectified images into 512×640. Meanwhile, we implement our DA-MVSNet in three stages with 1/4, 1/2, and original input images, respectively. From low-resolution to high-resolution stages, the number of depth hypothesis planes is 64, 32, and 8. Their corresponding depth intervals are set to 4, 2, and 1. The number of input images $N$ is set to 5. Our model is trained for 16 epochs with Adam optimizer [49]. The initial learning rate is 0.001, which is multiplied by 0.5 after 10, 12, and 14 epochs. As for BlendedMVS dataset, we train for 10 epochs with an initial learning rate of 0.0002, which is down-scaled by a factor of 2 after 6 and 8 epochs. During finetuning, the number of input images is 10, with the original resolution of 576×768. The batch size is 4 on one NVIDIA A6000 for the DTU and BlendedMVS datasets.

When testing on DTU, the image resolution is 864×1152, and the number of input images $N$ is set to 5. Besides, the number of depth hypothesis planes is the same as in the training process. As for the T&T dataset, the resolution of input images is either 1024×1920 or 1024×2048. The number of input images is 11. To evaluate the model on the DTU and T&T datasets, we use NVIDIA RTX3090 with 24G RAM.

### 4.2 Comparison experiments

#### 4.2.1 Results on DTU

We compare our results with the classical traditional method, COLMAP [2], and recent learning-based methods based on MVSNet. CasMVSNet [10] and TransMVSNet [32] are the typical ones representing the state-of-the-art methods in this field over the past two years, and the authors have opened their codes online. Therefore, we take these two methods as baselines and try our edge-aware modules separately. The quantitative results are shown in Table 1, which are calculated only considering the foreground regions of objects. The CasMVSNet and TransMVSNet improved by our modules show great enhancement in performance, where our overall metrics even rank first. These results imply that our plug-and-play modules can significantly improve the MVSNet-based methods having similar structures.

Table 1. Quantitative results of reconstructed point clouds on DTU.

| Method | Acc. (mm)↓ | Comp. (mm)↓ | Overall. (mm)↓ |
|---|---|---|---|
| Gipuma [1] | 0.283 | 0.873 | 0.578 |
| COLMAP [2] | 0.400 | 0.663 | 0.532 |
| CasMVSNet [10] | 0.336 | 0.403 | 0.370 |
| EPP-MVSNet [50] | 0.413 | 0.296 | 0.355 |
| RayMVSNet [20] | 0.341 | 0.319 | 0.330 |
| TransMVSNet [32] | 0.369 | 0.270 | 0.320 |
| VisMVSNet [30] | 0.369 | 0.361 | 0.365 |
| N2MVSNet [51] | 0.336 | 0.295 | 0.316 |
| CasMVSNet + ours | 0.333 | 0.308 | 0.320 |
| TransMVSNet + ours | 0.316 | 0.299 | 0.308 |

To illustrate the effect on the details of our modules, we also visualize the results in Figs. 5 and 6. The depth maps generated by the original method suffer from large errors in the regions of edges and details. In contrast, the overall error in our depth maps is substantially reduced, and the depth of edges and thin structures is also recovered much better than other methods, resulting in our depth map aligning with the RGB image better. In Fig. 6, the reconstructed point clouds display the details much clearer, as in red marks.

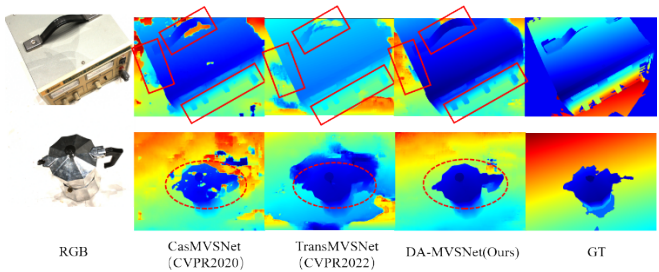

**Fig. 5.** Qualitative comparisons of estimated depth maps for Scan11 and Scan77 in DTU.

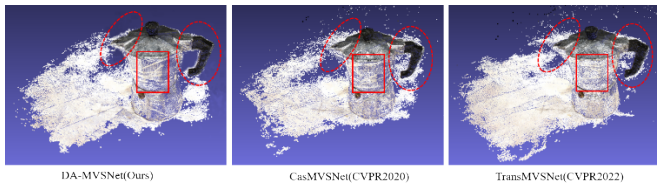

**Fig. 6.** Qualitative results on the reconstructed point cloud for Scan 77 in DTU.

### 4.2.2 Results on T&T

We further validate the generalization capability of our method on the T&T dataset. As introduced above, the T&T dataset includes two subsets, the intermediate and advanced, and each consists of several scenes. Consequently, experiments were conducted on both the subsets. The quantitative results are listed in Tables 2 and 3 separately. The F-score metric is obtained by the official website [47].

The experimental results also demonstrate that our module enhances the overall performance of the baseline networks. We rank first for the overall metric among all submissions on the advanced subset, which proves our effect in various scenarios.

In Table 2, the N2MVSNet [51] method outperforms our approach in some scenarios. However, the source code for the N2MVSNet network is not publicly available. Our module has plug-and-play compatibility, thus, we believe that integrating our module into the N2MVSNet network will similarly improve the overall performance, as evidenced by the preceding experimental results.

### 4.3 Ablation Study

Table 4 shows the ablation results of our DA-MVSNet. Our baseline is CasMVSNet [10], and all experiments were conducted with the same hyperparameters.

**Effect of GDE Module.**

As evidenced by the results in Table 4, setting the GDE modules into the branches of the MVS reconstruction process not only preserves the geometric structural relationships between object surfaces but also enhances the robustness of feature matching. This ultimately leads to an improvement in the quality of MVS reconstruction, i.e., the metric Acc decreases from 0.336 to 0.331, the metric Comp increases from 0.403 to 0.363, and the metric Overall decreases from 0.369 to 0.347.

As shown in Fig. 7, the qualitative results further validate the effectiveness of the GDE module. This approach markedly improves the reconstruction quality in object boundaries and detail regions.

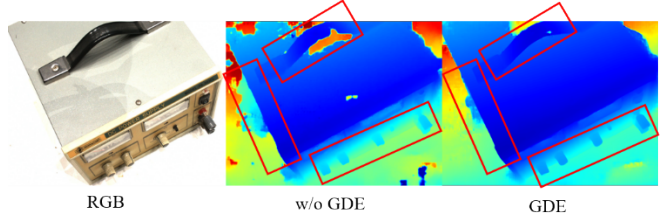

**Fig. 7.** Qualitative comparisons to evaluate the GDE module with Scan11 in DTU.

**Table 2.** Quantitative results on the intermediate subsets of T&T.

| Method | Intermediate (%) | | | | | | | | |
|---|---|---|---|---|---|---|---|---|---|
| | Mean↑ | Fam. | Fra. | Hor. | Lig. | M60. | Pan. | P.G. | Tra. |
| CasMVSNet [10] | 48.67 | 66.43 | 39.54 | 36.12 | 47.65 | 53.61 | 50.28 | 51.87 | 43.89 |
| EPP-MVSNet [50] | 61.68 | 77.86 | 60.54 | 52.96 | 62.33 | 61.69 | 60.34 | 62.44 | 55.30 |
| RayMVSNet [20] | 59.48 | 78.55 | 61.93 | 45.48 | 57.59 | 61.00 | 59.78 | 59.19 | 52.32 |
| TransMVSNet [32] | 61.24 | 77.51 | 60.83 | 51.32 | 61.24 | 62.36 | 59.85 | 59.50 | 57.34 |
| VisMVSNet [30] | 60.92 | 80.21 | 63.51 | 52.30 | 61.38 | 61.47 | 58.16 | 58.98 | 51.38 |
| N2MVSNet [51] | 62.14 | 80.39 | 65.64 | 51.08 | 62.33 | 62.30 | 61.89 | 59.02 | 54.47 |
| CasMVSNet + Ours | 56.93 | 73.19 | 57.80 | 39.73 | 62.02 | 57.13 | 54.61 | 58.59 | 52.36 |
| TransMVSNet + Ours | 61.81 | 77.97 | 61.41 | 56.96 | 63.48 | 61.32 | 59.62 | 58.59 | 55.11 |

**Table 3.** Quantitative results on the advanced subsets of T&T.

| Method | Advanced (%) | | | | | | |
|---|---|---|---|---|---|---|---|
| | Mean↑ | Aud. | Bal. | Cou. | Murs. | Pal. | Tem. |
| CasMVSNet [10] | 26.69 | 14.45 | 32.41 | 29.88 | 38.1 | 22.82 | 43.89 |
| EPP-MVSNet [50] | 35.72 | 21.28 | 39.74 | 35.34 | 49.21 | 30.00 | 38.75 |
| RayMVSNet [20] | - | - | - | - | - | - | - |
| TransMVSNet [32] | 37.00 | 24.84 | 44.59 | 34.77 | 46.49 | 34.69 | 36.62 |
| VisMVSNet [30] | 37.53 | 26.68 | 42.14 | 35.65 | 49.37 | 32.16 | 39.19 |
| N2MVSNet [51] | - | - | - | - | - | - | - |
| CasMVSNet + Ours | 37.95 | 25.62 | 43.54 | 38.03 | 45.54 | 33.57 | 41.43 |
| TransMVSNet + Ours | 39.27 | 28.54 | 44.23 | 38.92 | 48.79 | 34.74 | 40.38 |

Table 4. Ablation results with different components on the DTU evaluation dataset.

| | Module Settings | | | Mean Distance | | |
|---|---|---|---|---|---|---|
| | ADIA | IS Loss | GDE | Acc. (mm)↓ | Comp. (mm)↓ | Overall. (mm)↓ |
| (a) | | | | 0.336 | 0.403 | 0.369 |
| (b) | ✓ | | | 0.332 | 0.368 | 0.350 |
| (c) | | | ✓ | 0.331 | 0.363 | 0.347 |
| (d) | ✓ | | ✓ | 0.365 | 0.295 | 0.330 |
| (e) | ✓ | ✓ | | 0.325 | 0.367 | 0.346 |
| (f) | ✓ | ✓ | ✓ | 0.333 | 0.308 | 0.320 |

**Effect of ADIA Module.**

As shown in Table 4, the adaptive pixel-wise depth interval reallocated by our ADIA strategy can significantly decreases metric the Acc. from 0.336 to 0.332, and the Overall from 0.369 to 0.35 on DTU.

Figure 8 presents a qualitative comparison to evaluate the ADIA using two scans on the DTU benchmark. The depth maps estimated using the ADIA strategy exhibit superior overall quality compared to those obtained without it. The red boxes in the two scenes reveal that our method provides more accurate details.

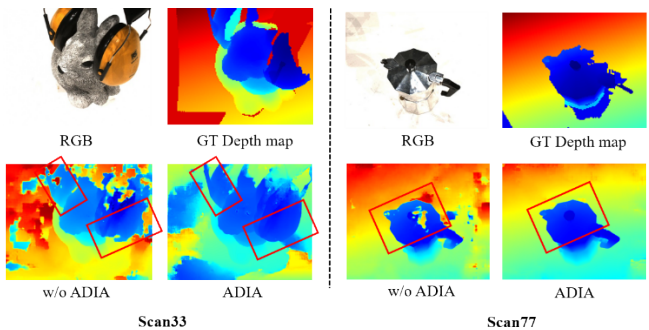

**Fig. 8.** Qualitative comparisons to evaluate the ADIA strategy with two scans.

As shown in Fig. 9, we visualize the configurations of the DHPs in the third stage for processing Scene 11 in the DTU dataset. With the ADIA strategy, the DHPs are set denser around the potential GT depth values and sparser far from it. Therefore, the final predicted depth value is closer to the ground truth.

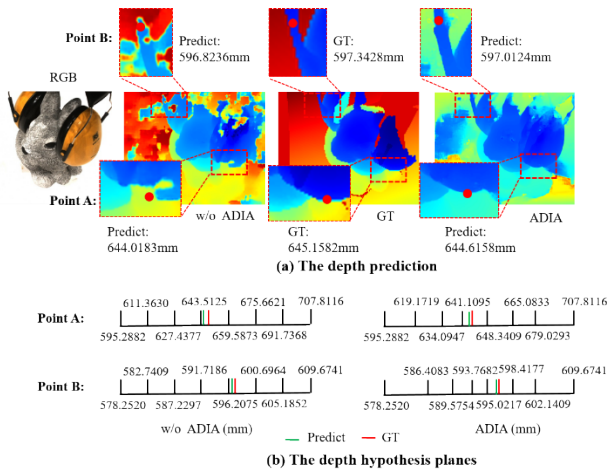

**Fig. 9.** Comparison for setting the DHPs.

## 5. Conclusion

In this paper, we propose a novel DA-MVSNet approach to improve the detailed regions. The proposed GDE module effectively maintains the geometric structural relationships between object surfaces during the reconstruction process and enhances the robustness of feature matching. The designed IS Loss amplifies the gradient flow in detailed regions to optimize the network training. The proposed ADIA strategy can further enhance the reconstruction precision. Extensive experiments demonstrate that our proposed method improves the quality in edge and detailed regions. These modules are plug-and-play and can be easily integrated into the existing MVSNet-based methods. In the future, we plan to embed geometric depth information into unsupervised or self-supervised MVS frameworks to improve generalization and accuracy performance while reducing memory consumption.

**Funding.** National Natural Science Foundation of China (61828501).

**Disclosures.** The authors declare no conflicts of interest.